%% file: acl_latex.tex
\pdfoutput=1

\PassOptionsToPackage{table,xcdraw}{xcolor}  

\documentclass[11pt]{article}

\usepackage[]{acl}

\usepackage{times}
\usepackage{latexsym}
\usepackage{multirow}
\usepackage{graphicx}
\usepackage[table,xcdraw]{xcolor}
\usepackage{caption}
\usepackage{subcaption}
\DeclareCaptionLabelFormat{andtable}{#1~#2  \&  \tablename~\thetable}

\usepackage{todonotes}
\usepackage[T1]{fontenc}

\usepackage[utf8]{inputenc}

\usepackage{microtype}

\usepackage{supertabular,booktabs}
\usepackage{xcolor}
\usepackage{color, colortbl,soul}
\definecolor{LGray}{gray}{0.9}
\definecolor{Gray}{gray}{0.8}
\definecolor{DGray}{gray}{0.7}
\sethlcolor{DGray}

\newcommand{\squishlist}{
 \begin{list}{$\bullet$}
  { \setlength{\itemsep}{0pt}
     \setlength{\parsep}{3pt}
     \setlength{\topsep}{3pt}
     \setlength{\partopsep}{0pt}
     \setlength{\leftmargin}{1.5em}
     \setlength{\labelwidth}{1em}
     \setlength{\labelsep}{0.5em} } }
\newcommand{\squishend}{
     \end{list}}
\title{Identifying Moments of Change from Longitudinal User Text}

\author{
\textbf{Adam Tsakalidis$^{1,2}$, Federico Nanni$^2$, Anthony Hills$^1$,}\\\textbf{Jenny Chim$^1$, Jiayu Song$^1$, Maria Liakata$^{1,2,3}$}\vspace{.2cm}\\
       $^1$ Queen Mary University of London, London, United Kingdom\\
       $^2$ The Alan Turing Institute, London, United Kingdom\\
       $^3$ University of Warwick, Coventry, United Kingdom\\
      \tt  a.tsakalidis;m.liakata@qmul.ac.uk\\
}

\begin{document}
\maketitle
\begin{abstract}
Identifying changes in individuals' behaviour and mood, as observed via content shared on online platforms, is increasingly gaining importance. Most research to-date on this topic focuses on either: (a) identifying individuals at risk or with a certain mental health condition given a batch of posts or (b) providing equivalent labels at the post level. A disadvantage of such work is the lack of a strong temporal component and the inability to make longitudinal assessments following an individual's trajectory and allowing timely interventions. Here we define a new task, that of identifying moments of change in individuals on the basis of their shared content online. The changes we consider are sudden shifts in mood (switches) or gradual mood progression (escalations). We have created detailed guidelines for capturing moments of change and a corpus of 500 manually annotated user timelines (18.7K posts). We have developed a variety of baseline models drawing inspiration from related tasks and show that the best performance is obtained through context aware sequential modelling. We also introduce new metrics for capturing rare events in temporal windows.

\end{abstract}

\input{1introduction}
\input{2related}
\input{3dataset}

\input{4experiments}

\input{5results}

\input{6conclusion}

\section{Ethics Statement}\label{sec:ethics}
Ethics institutional review board (IRB) approval was obtained from the corresponding ethics board of the University of Warwick prior to engaging in this research study. 
Our work involves ethical considerations around the analysis of user generated content shared on a peer support network (TalkLife). A license was obtained to work with the user data from TalkLife and a project proposal was submitted to them in order to embark on the project. 
The current paper focuses on the identification of moments of change (MoC) on the basis of content shared by individuals. These changes involve recognising sudden shifts in mood (switches or escalations).
Annotators were given contracts and paid fairly in line with University payscales. They were alerted about potentially encountering disturbing content and were advised to take breaks. The annotations are used to train and evaluate natural language processing models for recognising moments of change as described in our detailed guidelines.
Working with datasets such as TalkLife and data on online platforms where individuals disclose personal information involves ethical considerations~\cite{10.1145/2046556.2046558, kekulluoglu2020analysing}. Such considerations include careful analysis and data sharing policies to protect sensitive personal information. 
The data has been de-identified both at the time of sharing by TalkLife but also by the research team to make sure that no user handles and names are visible. Any examples used in the paper are either paraphrased or artificial.
Potential risks from the application of our work in being able to identify moments of change in individuals' timelines are akin to those in earlier work on personal event identification from social media and the detection of suicidal ideation. Potential mitigation strategies include restricting access to the code base and annotation labels used for evaluation. 

\paragraph{Limitations} Our work in this paper considers moments of change as changes in an individual's mood judged on the basis of their self-disclosure of their well-being. This is faced by two limiting factors: (a) users may not be self-disclosing important aspects of their daily lives and (b) other types of changes related to their mental health (other than their mood/emotions, such as important life events, symptoms etc.) may be taking place. Though our models could be tested in cases of non-self-disclosure (given the appropriate ground truth labels), the analysis and results presented in this work should not be used to infer any conclusion on such cases. The same also holds for other types of `moments of change' mentioned in \S\ref{sec:related} (e.g., transition to suicidal thoughts), as well as other types of changes, such as changes in an individual in terms of discussing more about the future, studied in \citet{althoff2016large}, or changes in their self-focus \cite{pyszczynski1987self} over time, which we do not examine in this current work.


\section*{Acknowledgements}
This work was supported by a UKRI/EPSRC Turing AI Fellowship to Maria Liakata (grant EP/V030302/1) and the Alan Turing Institute (grant EP/N510129/1). The authors would like to thank Dana Atzil-Slonim, Elena Kochkina, the anonymous reviewers and the meta-reviewer for their valuable feedback on our work, as well as the three annotators for their invaluable efforts in generating the longitudinal dataset.

\bibliographystyle{acl_natbib}
\bibliography{anthology,custom}

\input{7appendix}
\end{document}

%% file: 1introduction.tex
\section{Introduction}
Linguistic and other content from social media data has been used in a number of different studies to obtain biomarkers for mental health. This is gaining importance given the global increase in mental health disorders, the limited access to support services and the prioritisation of mental health as an area by the \citet{WHO2019}.  Studies using linguistic data for mental health focus on recognising specific conditions related to mental health (e.g., depression, bipolar disorder) \cite{husseini-orabi-etal-2018-deep}, or identifying self-harm ideation in user posts \cite{yates-etal-2017-depression,zirikly2019clpsych}. However, none of these works, even when incorporating a notion of time \cite{lynn2018clpsych,losada2020overview}, identify how an individual's mental health changes over time.
Yet being able to make assessments on a longitudinal level from linguistic and other digital content is important for clinical outcomes, and especially in mental health~\cite{velupillai2018using}.
The ability to detect changes in individual's mental health over time is also important in enabling platform moderators to prioritise interventions for vulnerable individuals 
\cite{wadden2021effect}. 
Users who currently engage with platforms and apps for mental health support ~\cite{neary2018state} would also benefit from being able to monitor their well-being in a longitudinal manner. 

\begin{figure}
    \centering
    \includegraphics[width=.95\linewidth]{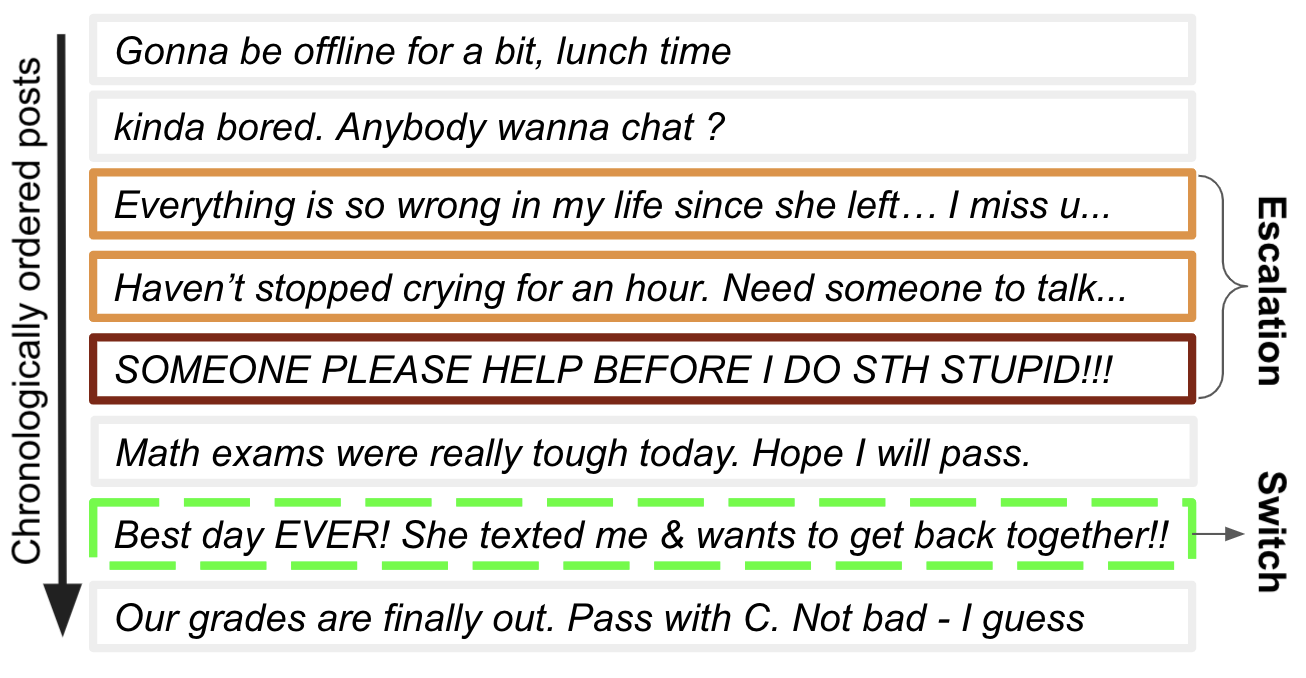}
    \caption{Example of an Escalation (with a darker ``peak'') and a Switch within a user's timeline.}
    \label{fig:introexample}
\end{figure}

Motivated by the lack of longitudinal approaches we introduce the task of \textit{identifying `Moments of Change' (MoC) from individuals' shared online content}. 
We focus in particular on two types of changes: \textit{Switches} -- mood shifts from positive to negative, or vice versa -- and \textit{Escalations} -- gradual mood progression 
(see Fig.~\ref{fig:introexample}, detailed in \S~\ref{sec:dataset}). 
Specifically we make the following contributions:

\squishlist
    \item We present the novel task of identifying moments of change in an individual's mood by analysing linguistic content shared online over time, along with a longitudinal dataset of 500 user timelines (18.7K posts, English language) from 500 users of an online platform.
    \item We propose a number of baseline models for automatically capturing  Switches/Escalations, inspired by sentence- and sequence-level state-of-the-art NLP approaches in related tasks. 
    \item 
    We introduce a range of temporally sensitive evaluation metrics for longitudinal NLP tasks adapted from the fields of change point detection \cite{van2020evaluation} and image segmentation \cite{arbelaez2010contour}. 
    \item We provide a thorough qualitative linguistic analysis of model performance. 
    
\squishend     

%% file: 2related.tex
\section{Related Work}\label{sec:related}

\paragraph{Social Media and Mental Health}
Online user-generated content provides a rich resource for computational modelling of wellbeing at both population and individual levels. Research has examined mental health conditions by analysing data from platforms such as Twitter and Reddit \cite{de2013predicting, coppersmith2014quantifying, cohan-etal-2018-smhd} as well as peer-support networks such as TalkLife \cite{10.1145/3290605.3300294}.  
Most such work relies on proxy signals for annotations (e.g., self-disclosure of diagnoses, posts on support networks) and is characterised by a lack of standardisation in terms of annotation and reporting practices \cite{chancellor2020methods}.
We have provided thorough annotation guidelines for Moments of Change that can aid mental health monitoring over time irrespective of the underlying condition.

\paragraph{Moments of Change (MoC)}
Little work has specifically focused on automatically capturing changes in user behaviour based on their social media posts. Within the health domain, \citet{guntuku2020variability} showed that a user's language on Facebook becomes more depressed and less informal prior to their visit to an emergency department. With respect to mental health, \citet{de_choudhury_discovering_2016} proposed to identify shifts to suicide ideation 
by predicting (or not) a transition from posting on a regular forum to a forum for suicide support. 
\citet{10.1145/3290605.3300294} examined moments of affective change in TalkLife users by identifying positive changes in sentiment at post-level with respect to a distressing topic earlier in a user's thread. In both cases MoC are overly specific and modelled through binary classification without any notion of temporal modelling.  

\paragraph{NLP for Mental Health}
More advanced NLP methods have been used for predicting psychiatric conditions from textual data, including self-harm, suicide ideation, eating disorders, and depression \cite{benton-etal-2017-multitask, kshirsagar2017detecting, yates-etal-2017-depression, husseini-orabi-etal-2018-deep, Jiang2020DetectionOM, shing-etal-2020-prioritization}. Researchers are increasingly adopting sequential modelling to capture temporal dynamics of language use and mental health. For example, \citet{cao-etal-2019-latent} encode microblog posts using suicide-oriented embeddings fed to an LSTM network to assess the suicidality risk at post level. \citet{sawhney-etal-2020-time,sawhney2021phase} improves further on predicting suicidality at post-level by jointly considering an emotion-oriented post representation and the user's emotional state as reflected through their posting history with temporally aware models. The recent shared tasks in eRisk also consider sequences of user posts in order to classify a user as a ``positive'' (wrt self-harm or pathological gambling) or ``control'' case \cite{losada2020overview,parapar2021overview}. While such work still operates at the post- or user-level it highlights the importance of temporally aware modelling. 

\paragraph{Related Temporal NLP Tasks}
Semantic change detection (SCD) aims to identify words whose meaning has changed over time. Given a set of word representations in two time periods, the dominant approach is to learn the optimal transformation using Orthogonal Procrustes~\cite{schonemann1966generalized} 
and measure the level of semantic change of each word via the cosine distance of the resulting vectors \cite{hamilton2016diachronic}. A drawback of this is the lack of connection between consecutive windows. \citet{tsakalidis-liakata-2020-sequential} addressed this through sequential modeling by encoding word embeddings in consecutive time windows and taking the cosine distance between future predicted and actual word vectors. Both approaches are considered as baselines for our task. 
First story detection (FSD) aims to detect new events reported in streams of textual data. Having emerged in the Information Retrieval community \cite{allan1998topic}, FSD has  been applied to streams of social media posts \cite{petrovic2010streaming}. FSD methods assume that a drastic change in the textual content of a document compared to previous documents signals the appearance of a new story. A baseline from FSD is considered in \S\ref{sec:approaches}.

%% file: 3dataset.tex
\section{Dataset creation} 
\label{sec:dataset}
We describe the creation of a dataset of individuals' \textit{timelines} annotated with Moments of Change. A \textit{user's timeline} $P_{s:e}^{(u)}$ is a subset of their history, a series of posts [$p_0$, ..., $p_n$] shared by user $u$ between dates $s$ and $e$. A \textit{``Moment of Change''} (MoC) is a particular point or period (range of time points) within $[s,e]$ where the behaviour or mental health status of an individual changes. While MoC can have different definitions in various settings, in this paper we are particularly interested in capturing MoC pertaining to an individual's mood. Other types of MoC can include life events, the onset of symptoms or turning points (e.g., moments of improvement, difficult moments or moments of intervention within therapy sessions).\footnote{A limitation of our work stems from the fact that MoC are revealed to us by the user's shared content (i.e., we cannot identify changes in a user's well-being unless these are expressed online). We provide details on the limitations of our work in the Ethics Statement (\S\ref{sec:ethics}).} We address two types of Moments of Change: \textbf{Switches} (sudden mood shifts from positive to negative, or vice versa) and \textbf{Escalations} (gradual mood progression from neutral or positive to more positive or neutral or negative to more negative). Capturing both sudden and gradual changes in individuals' mood over time is recognised as important for monitoring mental health conditions \cite{Lutz2013TheUA,SHALOM2020101827} and is one of the dimensions to measure in psychotherapy \cite{barkham2021bergin}.


\subsection{Data Acquisition}
Individual's timelines are extracted from  Talklife\footnote{\url{https://www.talklife.com/}}, a peer-to-peer network for mental health support. Talklife incorporates all the common features of social networks -- post sharing, reacting, commenting, etc. Importantly, it provides a rich resource for computational analysis of mental health~\cite{10.1145/3290605.3300294,sharma2020computational,saha2020causal} given that content posted by its users focuses on their daily lives and well-being.   

A complete collection between Aug'11--Aug'20 (12.3M posts, 1.1M users) 
was anonymised and provided to our research team in a secure environment upon signing a License Agreement. In this environment, 500 user timelines were extracted (\S\ref{sec:timeline_creation}) and an additional anonymisation step was performed to ensure that usernames were properly hashed when present in the text. The 500 timelines were subsequently annotated using our bespoke annotation tool (\S\ref{sec:annotations}) to derive the resulting longitudinal dataset (\S\ref{sec:finaldataset}).

\subsection{Timeline Extraction}\label{sec:timeline_creation}
    Existing work extracts user timelines either based on a pre-determined set of timestamps (e.g., considering the most recent posts by a user) \cite{sawhney-etal-2020-time} or by selecting a window of posts around mentions of specific phrases (e.g., around self-harm) \cite{mishra_snap-batnet_2019}. The latter introduces potential bias into subsequent linguistic analysis \cite{olteanu2019social}, while the former could result into selecting timelines from a particular time period -- hence potentially introducing temporally-dependent linguistic or topical bias (e.g., a focus on the COVID-19 pandemic). Here we instead extract timelines around points in time where a user's posting behaviour has changed. Our hypothesis is that such changes in a user's posting frequency could be indicative of changes in their lives and/or mental health. Such association between changes in posting behaviour on mental health fora and changes in mental health has been assumed in prior literature \cite{de_choudhury_discovering_2016}.

\paragraph{Identifying changes in posting frequency} We create a time series of each user's daily posting frequency based on their entire history. We then employ a change-point detection model to predict the intensity of daily post frequency by the given user. Bayesian Online Change-point Detection \cite{adams_bayesian_2007} with a Poisson-Gamma underlying predictive model \cite{bocpmd_pg_2018} was chosen as our model, due to its highly competitive performance 
\cite{van2020evaluation} and the fact that  extracted timelines using this method had the highest density of MoC compared to a number of different timeline extraction (anomaly detection and keyword-based) methods for the same dataset. 

\begin{figure}%
    \centering
    \subfloat[\centering Posts per Timeline]{{\includegraphics[width=.43\linewidth]{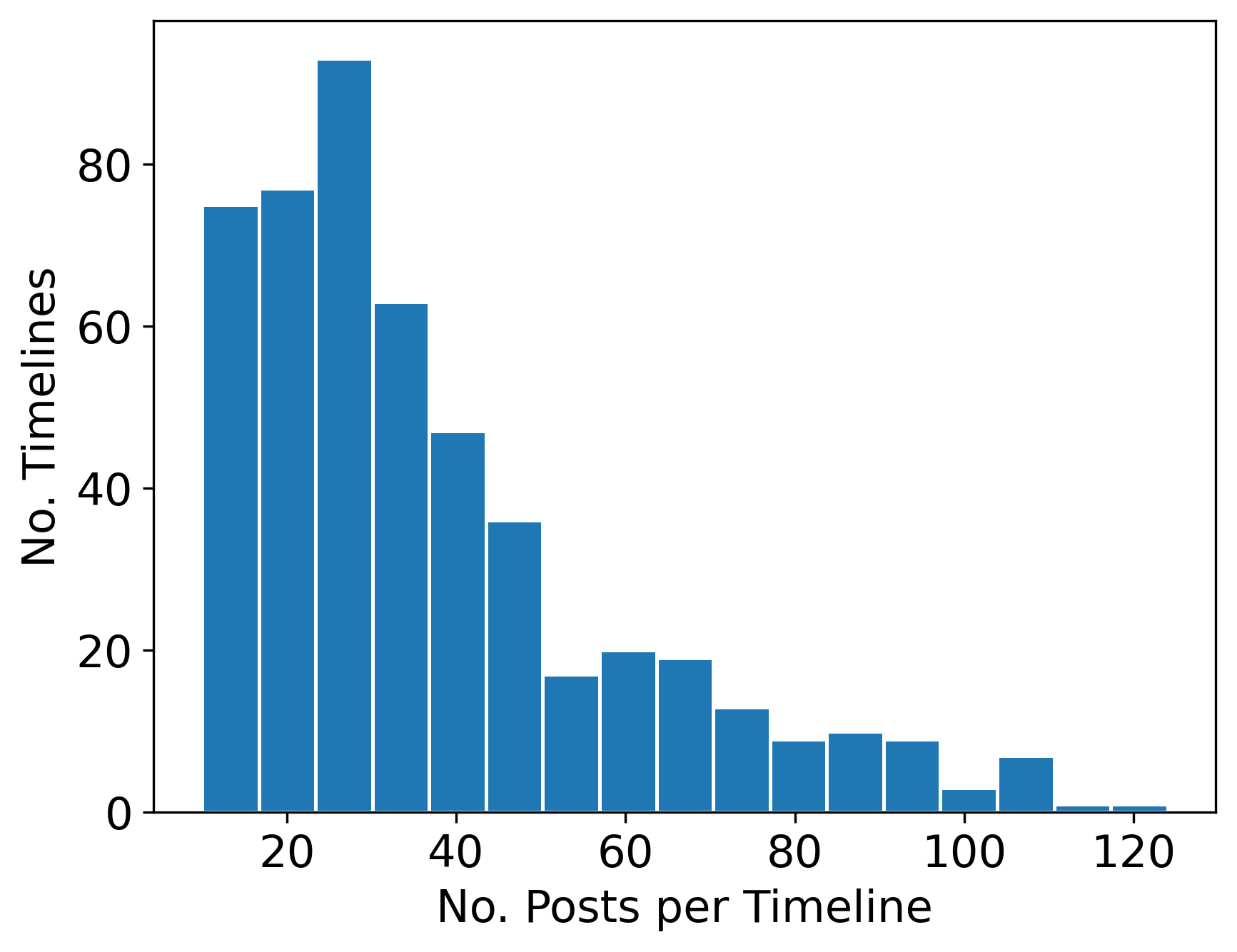} }}%
    \qquad
    \subfloat[\centering Posts per MoC Area]{{\includegraphics[width=.43\linewidth]{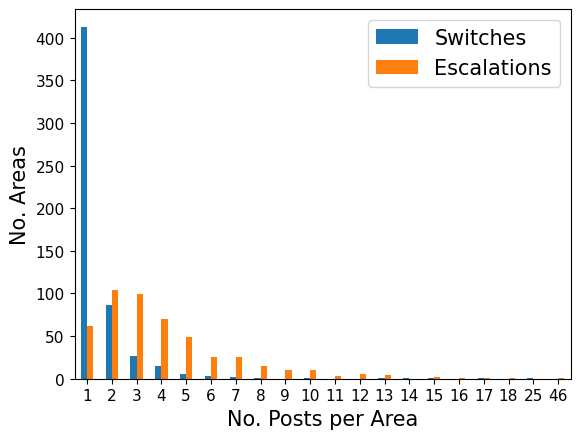} }}%
    \caption{Distributions in our dataset.}%
    \label{fig:combined}%
\end{figure}

\paragraph{Extracting timelines around change-points} Upon detecting candidate MoC as change-points in posting frequency, we generated candidate timelines for annotation by extracting all of the user's posts within a seven-day window around each change-point. We controlled for timeline length (between 10 and 150 posts, set empirically) so that they were long enough to enable annotators to notice a change but not so long as to hinder effective annotation. This control for timeline length means that our subsequent analysis is performed (and models are trained and evaluated) on time periods during which the users under study are quite active; however, the upper bound of 150 posts in 15 days set for each timeline also ensures that we do not bias (or limit) our analysis on extremely active users. 
Finally, to ensure linguistic diversity in our dataset, 500 timelines extracted in this way were chosen for annotation at random, each corresponding to a different individual. The resulting dataset consists of 
18,702 posts ($\mu$=35, $SD$=22 per timeline; range of timeline length=[10,124], see Fig.~\ref{fig:combined}(a)).

\subsection{Annotations of MoC}\label{sec:annotationtool} 

\begin{figure}
\centering
\includegraphics[width=\linewidth]{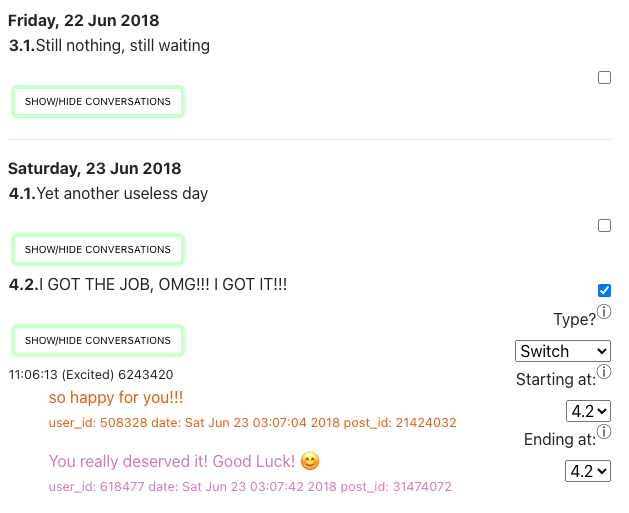}
\caption{Annotating a `Switch' on our interface (\S\ref{sec:annotationtool}).}
\label{fig:switch}
\end{figure}

\paragraph{Annotation Interface} An annotation interface was developed to allow efficient viewing and annotation of a timeline 
(see snippet in Fig.~\ref{fig:switch}). 
Each post in a timeline was accompanied by its timestamp, the user's self-assigned emotion and any associated comments (color-coded, to highlight recurrent users involved within the same timeline). Given the context of the entire timeline, annotations for \textit{MoC} are  performed at post level: if an annotator marks a post as a \textit{MoC}, then they specify whether it is (a) the beginning of a Switch or (b) the peak of an Escalation (i.e., the most positive/negative post of the Escalation). Finally, the range of posts pertaining to a MoC (i.e., all posts in the Switch/Escalation) need to be specified.

\paragraph{Data annotation}\label{sec:annotations}
After a round of annotations for guideline development with PhD students within the research group (co-authors of the paper), we recruited three external annotators to manually label the 500 timelines. They 
all have University degrees in humanities disciplines and come from three different countries; one of them is an English native speaker.
Annotators were provided with a set of annotation guidelines containing specific examples, which were enriched and extended during iterative rounds of annotation.\footnote{Guidelines are available at \url{https://github.com/Maria-Liakata-NLP-Group/Annotation-guidelines}.} 
Annotators completed 2 hands-on training sessions with a separate set of 10 timelines, where they were able to ask questions and discuss opinions to address cases of disagreement. Following the initial training phase, we performed spot checks to provide feedback and answer any questions while they labelled the full dataset (n=500 timelines). Annotators were encouraged to take breaks whenever needed, due to the nature of the content. On average, each annotator spent about 5 minutes on annotating a single timeline.


\subsection{Deriving the final gold standard}\label{sec:finaldataset}

\begin{table}
\centering
\resizebox{.95\linewidth}{!}{%
\begin{tabular}{ lcc } 
\hline
\textbf{Label} & \textbf{Perfect Agreement} & \textbf{Majority} \\ 
\hline
None (O) & 0.69 & 0.89  \\
Switch (IS) & 0.08 & 0.30 \\
Escalation (IE) & 0.19 & 0.50 \\

\end{tabular}%
}
\caption{Inter Annotator Agreement (IAA).}
\label{table:iaa}
\end{table}

The annotation of MoC is akin to assessment of anomaly detection methods since MoC (Switches and Escalations) are rare, with the majority of posts not being annotated (label `None'). 
Measuring the agreement in such settings is therefore complex, as established metrics such as Krippendorff’s Alpha and Fleiss' Kappa would generally yield a low score. This is due to the unrealistically high expected chance agreement \cite{feinstein1990high}, which cannot be mitigated by the fact that annotators do agree on the majority of the annotations (especially on the `None' class). For this reason, we have used as the main indicator 
the per label positive agreement computed as the ratio of the number of universally agreed-upon instances (the intersection of posts associated with that label) over the total number of instances (the union of posts associated with that label). As highlighted in Table~\ref{table:iaa}, while perfect agreement for `None' is at 69\%,  perfect agreement on Escalations and Switches is at 19\% and 8\%, respectively. However, if instead of perfect agreement we consider majority agreement  (where two out of three annotators agree), these numbers drastically increase (30\% for Switches and 50\% for Escalations). Moreover, by examining the systematic annotation preferences of our annotators we have observed that the native speaker marked almost double the amount of Switches compared to the other two annotators, in particular by spotting very subtle cases of mood change. We have thus decided to generate a gold standard based on majority decisions, comprising only cases where at least two out of three annotators agree with the presence of a MoC. The rare cases of complete disagreement have been labelled as `None'. We thus have 2,018 Escalations and 885 Switches from an overall of 18,702 posts (see Fig.~\ref{fig:combined}(b) for the associated lengths in \#posts). In future work we plan to consider aggregation methods based on all annotations or approaches for learning from multiple noisy annotations \cite{paun2021aggregating}.



%% file: 4experiments.tex
\section{Models \& Experiment Design}
Our aim is to detect and characterise the types of MoC based on a user's posting activity. We therefore treat this problem as a supervised classification task (both at post level and in a sequential/timeline-sensitive manner, as presented in~\S\ref{sec:approaches}) rather than an unsupervised task, even though we also consider effectively baselines with unsupervised components (FSD, SCD in~\S\ref{sec:approaches}). Contrary to traditional sentence or document-level NLP tasks, we incorporate timeline-sensitive evaluation metrics that account for the sequential nature of our model predictions (\S\ref{sec:evaluation}).

Given a user's timeline, the aim is to classify each post within it as belonging to a  \textit{``Switch''} (IS), an \textit{``Escalation''} (IE), or \textit{``None''} (O). At this point we don't distinguish between beginnings of switches/peaks of escalations and other posts in the respective ranges. While the task is sequential by definition, we train models operating both at the post level in isolation and sequential models at the timeline-level (i.e., accounting for user's posts over time), 
as detailed in \S\ref{sec:approaches}. We contrast model performance using common post-level classification metrics as well as novel timeline-level evaluation approaches  (\S\ref{sec:evaluation}). This allows us to investigate the impact of (a) accounting for severe class imbalance and (b) longitudinal modelling. We have randomly divided the annotated dataset into 5 folds (each containing posts from 100 timelines) to allow reporting results on all of the data through cross-validation.

\subsection{Evaluation Settings}
\label{sec:evaluation}
\begin{figure*}
\centering
\begin{subfigure}{.5\textwidth}
  \centering
  \includegraphics[width=.9\textwidth]{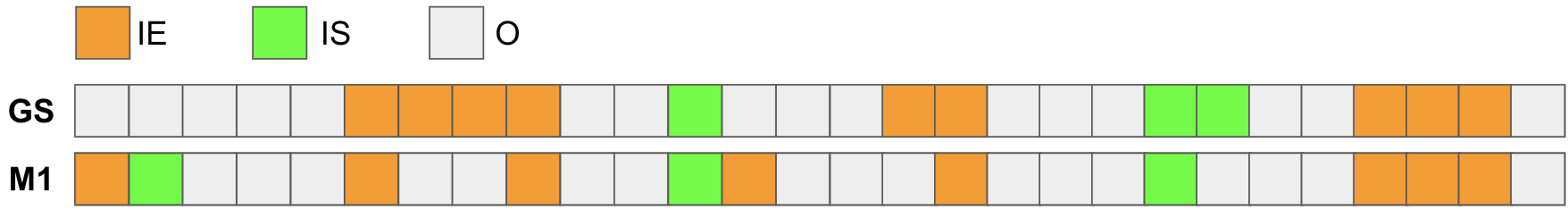}
\end{subfigure}%
\begin{subfigure}{.5\textwidth}
  \centering
  \includegraphics[width=.9\textwidth]{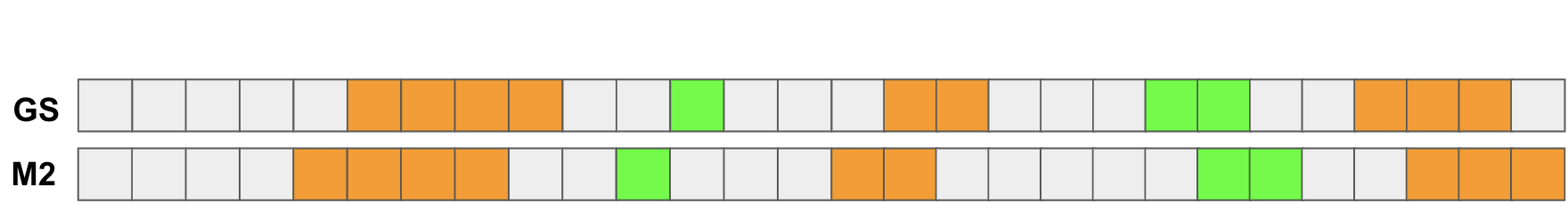}
\end{subfigure}
\caption{Actual (\textit{GS}, shown twice) \textit{vs} Predicted labels for each post (square) of a single timeline, by two models (\textit{M1}, \textit{M2}). Although \textit{M2} provides a more faithful `reconstruction' of the user's mood over time (the predictions are identical but shifted slightly in time), all post-level evaluation metrics for \textit{M1} are greater or equal to those obtained by \textit{M2} for the two minority classes (IE and IS).}
\label{fig:evalexample}
\end{figure*}

\paragraph{Post-level} We first assess model performance on the basis of standard evaluation metrics at the post level (Precision, Recall, F1 score). These are obtained per class and macro-averaged, to better emphasize performance in the two minority class labels (IS \& IE). However, post-level metrics are unable to show: (a) the expected accuracy at the timeline level (see example in Fig.~\ref{fig:evalexample}) and (b) model suitability in predicting \textit{regions} of change. 
These aspects are particularly important since we aim to build models capturing MoC over time.

\paragraph{Timeline-level} Our first set of timeline-level evaluation metrics are inspired from work in change-point detection \cite{van2020evaluation} and mirror the post-level ones, albeit operating on a window and timeline basis. Specifically, working on each timeline and label type independently, we calculate Recall $R_w^{(l)}$ (Precision $P_w^{(l)}$) by counting as ``correct'' a model prediction for label $l$ if the prediction falls within a window of $w$ posts around post labelled $l$ in the gold standard. Formally:\vspace{.08cm}

\begin{center}
    $R_w^{(l)} = \frac{|TP_w(M^{(l)}, GS^{(l)})|}{|GS^{(l)}|},
    P_w^{(l)} = \frac{|TP_w(M^{(l)}, GS^{(l)})|}{|M^{(l)}|}$,
\end{center}

\vspace{.08cm}\noindent where $TP_w$ denotes the true positives that fall within a range of $w$ posts and $M^{(l)}$/$GS^{(l)}$ are the predicted/actual labels for $l$, respectively. Note that each prediction can only be counted once as ``correct''. $R_w^{(l)}$ and $P_w^{(l)}$ are calculated on every timeline and are then macro-averaged. 

The second set of our timeline-level evaluation metrics is adapted from the field of image segmentation \cite{arbelaez2010contour}. Here we aim at evaluating model performance based on its ability to capture \textit{regions of change} (e.g., in Fig~\ref{fig:evalexample}, `GS' shows a timeline with three (two) such regions of Escalations (Switches)). For each such true region $R^{(l)}_{GS}$, we define its overlap $O(R_{GS}^{(l)}, R_M^{(l)})$ with each predicted region $R^{(l)}_M$ as the intersection over union between the two sets. This way, we can get recall and precision oriented \textit{coverage} metrics as follows:
\begin{center}
    \resizebox{.48\textwidth}{!}{$C_r^{(l)} (M\rightarrow GS) = \frac{1}{\sum_{R_{GS}^{(l)}}{|R_{GS}^{(l)}}|} \sum_{R_{GS}^{(l)}}      {|R_{GS}^{(l)}| \cdot max_{R_M^{(l)}} \{O(R_{GS}^{(l)}, R_M^{(l)})\}}$},
\resizebox{.48\textwidth}{!}{$C_p^{(l)} (M\rightarrow GS) = \frac{1}{\sum_{R_{M}^{(l)}}{|R_{M}^{(l)}}|} \sum_{R_{M}^{(l)}}      {|R_{M}^{(l)}| \cdot max_{R_{GS}^{(l)}} \{O(R_{GS}^{(l)}, R_{M}^{(l)})\}}$}.
  \end{center}  
\vspace{.08cm}The coverage metrics are calculated on the timeline basis and macro-averaged similarly to $R_w^{(l)}$ and $P_w^{(l)}$. Using a set of evaluation metrics, each capturing a different aspect of the task, ensures assess to model performance from many different angles.

\subsection{Baseline Models}\label{sec:approaches}
We have considered different approaches to addressing our task: 

\vspace{.2cm}
\noindent\textbf{(i) Na\"ive} methods, specifically a Majority classifier (predicting always ``None'') and a ``Random'' predictor, picking a label based on the overall label distribution in the dataset. It has been shown that comparisons against such simple baselines is essential to assess performance in computational approaches to mental health~\cite{tsakalidis2018can}. 

\vspace{.2cm}
\noindent\textbf{(ii) Post-level} supervised models operating on posts in isolation (i.e., ignoring post sequence in a user's timeline): (a) Random Forest \cite{breiman2001random} on tfidf post representations (\texttt{RF-tfidf}); (b) BiLSTM \cite{huang2015bidirectional} operating on sequences of word embeddings (\texttt{BiLSTM-we});
  (c) \texttt{BERT(ce)} \cite{devlin-etal-2019-bert} using the cross-entropy loss; and (d) \texttt{BERT(f)} trained using the alpha-weighted focal loss \cite{lin2017focal}, which is more appropriate for 
 imbalanced datasets.

\vspace{.2cm}
\noindent\textbf{(iii) Emotion Classification} 
We used DeepMoji (\texttt{EM-DM})~\cite{felbo2017} and Twitter-roBERTa-base (\texttt{EM-TR}) from TweetEval '20 ~\cite{barbieri-etal-2020-tweeteval} operating on the post-level, to generate softmax probabilities for each emotion (64 for \texttt{EM-DM}, 4 for \texttt{EM-TR}). These provide meta-features to a BiLSTM to obtain timeline-sensitive models for identifying MoC.

\vspace{.2cm}
\noindent\textbf{(iv) First Story Detection} \texttt{(FSD)}. We have used two common approaches for comparing a post to the \emph{n} previous ones: representing the previous posts as (i) a single centroid or (ii) the nearest neighbour to the current post among them \cite{allan1998topic,petrovic2010streaming}. In both cases, we calculate the cosine similarity of the current and previous posts. The scores are then fed into a BiLSTM as meta-features for a sequential model. Results are reported for the best method only.

\vspace{.2cm}
\noindent\textbf{(v) Semantic Change Detection (SCD)}. Instead of the standard task of comparing word representations in consecutive time windows, we consider a user being represented via their posts at particular points in time. 
We follow two approaches. The first is an Orthogonal Procrustes approach \cite{schonemann1966generalized} operating on post vectors (\texttt{SCD-OP}). Our aim here is to find the optimal transformation across  consecutive representations, with higher errors 
being indicative of a change in the user's behaviour. 
In the second approach (\texttt{SCD-FP}) a BiLSTM is trained on the user's $k$ previous posts in order to predict the next one \cite{tsakalidis-liakata-2020-sequential}. Errors in prediction are taken to signal changes in the user. In both cases, we calculate the dimension-wise difference between the actual and the transformed/predicted representations (post vectors) and use this as a meta-feature to a BiLSTM to obtain a time-sensitive model.

\vspace{.2cm}
\noindent\textbf{(vi) Timeline-sensitive}. From our (ii) post-level classifiers, \texttt{BERT(f)} tackles the problem of imbalanced data but fails to model the task in a longitudinal manner. To remedy this, we employ \texttt{BiLSTM-bert}, which treats a timeline as a sequence of posts to be modelled, each being represented via the \texttt{[CLS]} representation of \texttt{BERT(f)}.
To convert the post-level scores/representations from (iii)-(v) above into time-sensitive models we used the same BiLSTM from (vi), operating at the timeline-level. Details for each model and associated hyperparameters are in the Appendix.


%% file: 5results.tex
\begin{table*}[h]
\centering
\resizebox{.99\textwidth}{!}{%
\begin{tabular}{|ll|rrr|rrr|rrr|rrr||rr|rr|rr|rr|}
\hline
& &\multicolumn{12}{c||}{\textbf{Post-level Evaluation}} & \multicolumn{8}{c|}{\textbf{Coverage-based Metrics}} \\
 &
  \textbf{}& 
  \multicolumn{3}{c|}{\textbf{IS}} &
  \multicolumn{3}{c|}{\textbf{IE}} &
  \multicolumn{3}{c|}{\textbf{O}} &
  \multicolumn{3}{c||}{\textbf{macro-avg}} &\multicolumn{2}{c|}{\textbf{IS}} &
  \multicolumn{2}{c|}{\textbf{IE}} &
  \multicolumn{2}{c|}{\textbf{O}} &
  \multicolumn{2}{c|}{\textbf{macro-avg}} \\ \cline{3-22}
                   & \textbf{}            & P & R & F1 & P & R & F1 & P & R & F1 & P & R & F1 & $C_p$&$C_r$&$C_p$&$C_r$&$C_p$&$C_r$&$C_p$&$C_r$ \\ \hline
          \parbox[t]{2mm}{\multirow{2}{*}{\rotatebox[origin=c]{90}{Na\"ive}}}
        & \cellcolor{LGray}\textbf{Majority}    & \cellcolor{LGray}--  & \cellcolor{LGray}.000  & \cellcolor{LGray}.000   & \cellcolor{LGray}--  & \cellcolor{LGray}.000  & \cellcolor{LGray}.000   &   \cellcolor{LGray}.845 & \cellcolor{LGray}1.000 & \cellcolor{LGray}.916  &  \cellcolor{LGray}.282 & \cellcolor{LGray}.333 & \cellcolor{LGray}.305  &  \cellcolor{LGray}--  & \cellcolor{LGray}.000&   \cellcolor{LGray}--  & \cellcolor{LGray}.000&\cellcolor{LGray}.619 & \cellcolor{LGray}.559& \cellcolor{LGray}.206& \cellcolor{LGray}.186\\
\multirow{-2}{*}{} & \textbf{Random}      &   .047&  .047 &  .047  &   .108   & .108 & .108 & .845  & .845   &  .845   & .333   &  .333 & .333   &.031&.045&.033&.096&.386&.452&.150&.198 \\ \hline

\parbox[t]{2mm}{\multirow{4}{*}{\rotatebox[origin=c]{90}{Post-level}}}                   &  \cellcolor{LGray}\textbf{RF-tfidf}    &  \cellcolor{LGray}.294 & \cellcolor{LGray}.006  & \cellcolor{LGray}.011   &  \cellcolor{LGray}\textbf{.568} & \cellcolor{LGray}.087 & \cellcolor{LGray}.151 & \cellcolor{LGray}.852&\cellcolor{DGray}\textbf{.991}&\cellcolor{DGray}\textbf{.917}&\cellcolor{LGray}.571 &\cellcolor{LGray}.361& \cellcolor{LGray}.360&\cellcolor{LGray}.250&\cellcolor{LGray}.005&\cellcolor{LGray}.152&\cellcolor{LGray}.087&\cellcolor{LGray}.632&\cellcolor{LGray}.602&\cellcolor{LGray}.345&\cellcolor{LGray}.231 \\
                   & \textbf{BiLSTM-we} & .245  & .119  &  .160  & .416  & .347  & .378   & .878  & .923  &  .900  &  .513 &.463 &.479 &.173&.091&.138&.330&.557&.606&.289&.342 \\
                   &\cellcolor{LGray}\textbf{BERT(ce)}   & \cellcolor{LGray}.285  & \cellcolor{LGray}.186  &  \cellcolor{LGray}.222  &  \cellcolor{LGray}.454 &	\cellcolor{LGray}.368 &	\cellcolor{LGray}.406  &  \cellcolor{LGray}.883 &	\cellcolor{LGray}.921&	\cellcolor{LGray}.901 &    \cellcolor{LGray}.540&	\cellcolor{LGray}.492&	\cellcolor{LGray}.510 &\cellcolor{LGray}.247&\cellcolor{LGray}.163&\cellcolor{LGray}.172&\cellcolor{LGray}\textbf{.344}&\cellcolor{LGray}.578&\cellcolor{LGray}\textbf{.621}&\cellcolor{LGray}.332&\cellcolor{LGray}.376 \\
\multirow{-4}{*}{} & \textbf{BERT(f)}     &   .260&	\cellcolor{DGray}\textbf{.321}&	\textbf{.287}& .401 &	\cellcolor{DGray}\textbf{.478}&	.436  &  \textbf{.898}	& .864&	.881  & .520&	 \cellcolor{DGray}\textbf{.554}&	\textbf{.534}  &.227&\cellcolor{DGray}\textbf{.269}&.160&\cellcolor{DGray}\textbf{.423}&.503&.567&.297&\cellcolor{DGray}\textbf{.420}   \\ \hline

\parbox[t]{2mm}{\multirow{6}{*}{\rotatebox[origin=c]{90}{Timeline-level}}}  

         & \cellcolor{LGray}\textbf{FSD}    & \cellcolor{LGray}--  & \cellcolor{LGray}.000  &\cellcolor{LGray} .000   &\cellcolor{LGray} --  &\cellcolor{LGray} .000  &\cellcolor{LGray} .000   &\cellcolor{LGray}   .845 &\cellcolor{LGray} 1.000 &\cellcolor{LGray} .916  & \cellcolor{LGray} .282 &\cellcolor{LGray} .333 & \cellcolor{LGray}.305  & \cellcolor{LGray} --  &\cellcolor{LGray}.000& \cellcolor{LGray}  --  &\cellcolor{LGray}.000&\cellcolor{LGray}.619 &\cellcolor{LGray}.559&\cellcolor{LGray}.206&\cellcolor{LGray}.186\\

                   & \textbf{EM-TR}       & .344  & .036  & .065   & .444  & .248  &  .318  & .865   & \textbf{.957}  & .909   &   .551 &.414& .431  &.297&.024&.273&.104&.639&.589&.403&.239 \\
                   
                 & \cellcolor{LGray}\textbf{EM-DM}       & \cellcolor{DGray}\textbf{.533}  & \cellcolor{LGray}.118  &\cellcolor{LGray} .193   &\cellcolor{LGray} .479  &\cellcolor{LGray} .351  & \cellcolor{LGray} .405  &\cellcolor{LGray} .880  & \cellcolor{LGray}.948  &\cellcolor{LGray} .913   &  \cellcolor{DGray}\textbf{.631} & \cellcolor{LGray}.472 &\cellcolor{LGray} .504  &\cellcolor{DGray}\textbf{.347}&\cellcolor{LGray}.023&\cellcolor{DGray}\textbf{.363}&\cellcolor{LGray}.177&\cellcolor{LGray}.646&\cellcolor{LGray}.592&\cellcolor{DGray}\textbf{.452}&\cellcolor{LGray}.264  \\

                   &\textbf{SCD-OP}      &  .200 & .005  &  .009  & .478  & .408  & .440   & .882  & .947  &  .913  & .520 & .453  &  .454 &.167&.001&.344&.180&\textbf{.663}&.609&.391&.263  \\
                   
                   & \cellcolor{LGray}\textbf{SCD-FP}      & \cellcolor{LGray} .270 &\cellcolor{LGray} .082  &  \cellcolor{LGray}.126  & \cellcolor{LGray} .503 &\cellcolor{LGray} .370  & \cellcolor{LGray}.426   & \cellcolor{LGray} .880 & \cellcolor{LGray}.944  & \cellcolor{LGray} .911  &\cellcolor{LGray} .551  &\cellcolor{LGray} .465  & \cellcolor{LGray}.488 &\cellcolor{LGray}.227&\cellcolor{LGray}.039&\cellcolor{LGray}.317&\cellcolor{LGray}.254&\cellcolor{LGray}.649&\cellcolor{LGray}.611&\cellcolor{LGray}.398&\cellcolor{LGray}.301  \\
\multirow{-6}{*}{} & \textbf{BiLSTM-bert} &    \textbf{.397}	&\textbf{.264}&	\cellcolor{DGray}\textbf{.316}   &  \cellcolor{DGray}\textbf{.568} &	\textbf{.461}&	\cellcolor{DGray}\textbf{.508} & \textbf{.898}	&.936	&\textbf{.917}   &  \textbf{.621}	& \textbf{.553}	& \cellcolor{DGray}\textbf{.580} &\textbf{.331}&\textbf{.197}&\textbf{.345}&.340&\cellcolor{DGray}\textbf{.664}&\cellcolor{DGray}\textbf{.656}&\textbf{.447}&\textbf{.398}  \\ \cline{1-22} 
\end{tabular}%
}
\caption{Post-level and Coverage-based evaluation for each model (\hl{\textbf{first}} and \textbf{second} highest scores are highlighted).}
\label{tab:postlevelresults}
\end{table*}

\begin{figure*}[ht]
\begin{subfigure}{.245\textwidth}
  \centering
  \includegraphics[width=\textwidth]{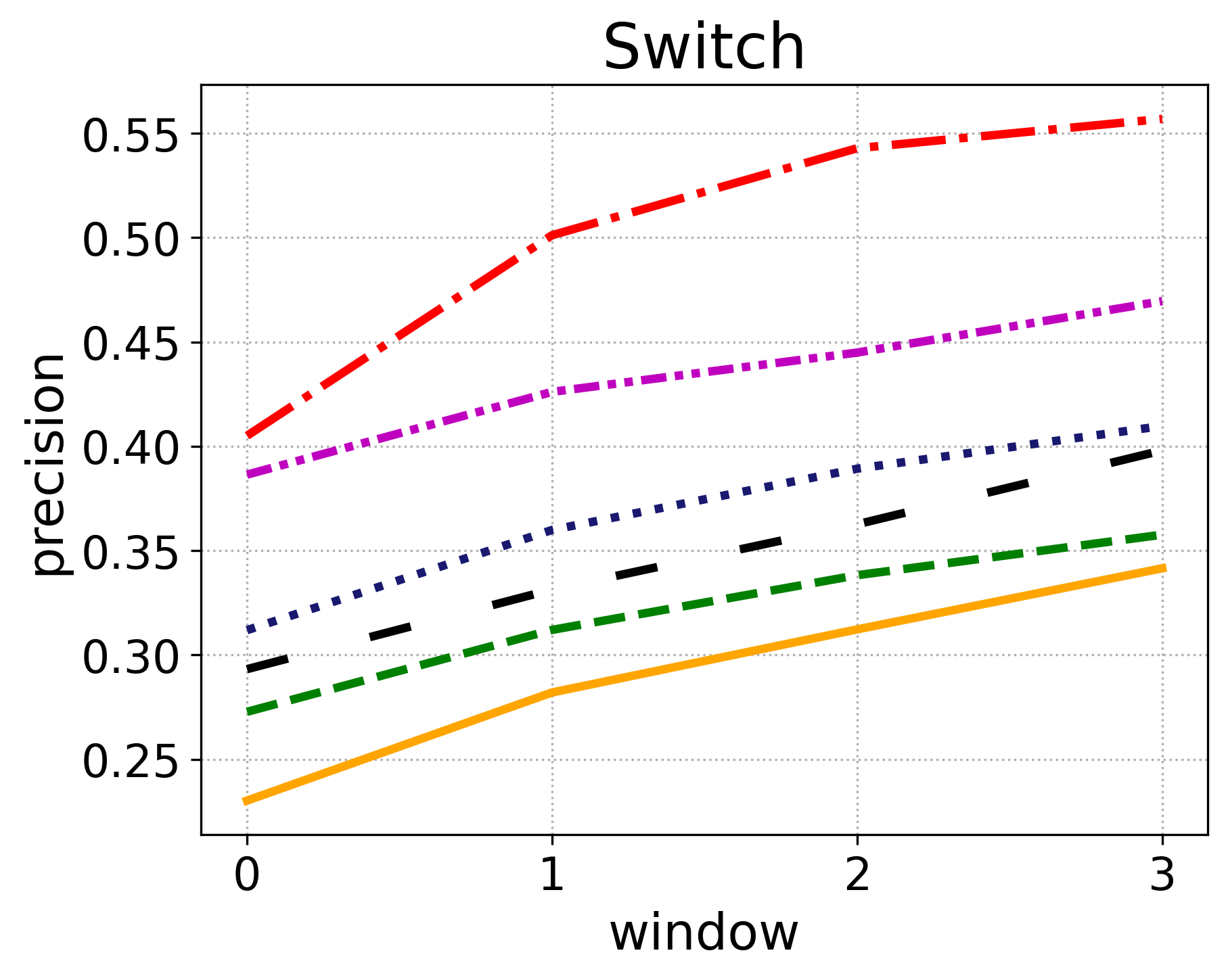}  
  \label{fig:sub-first}
\end{subfigure}
\begin{subfigure}{.245\linewidth}
  \centering
  \includegraphics[width=\linewidth]{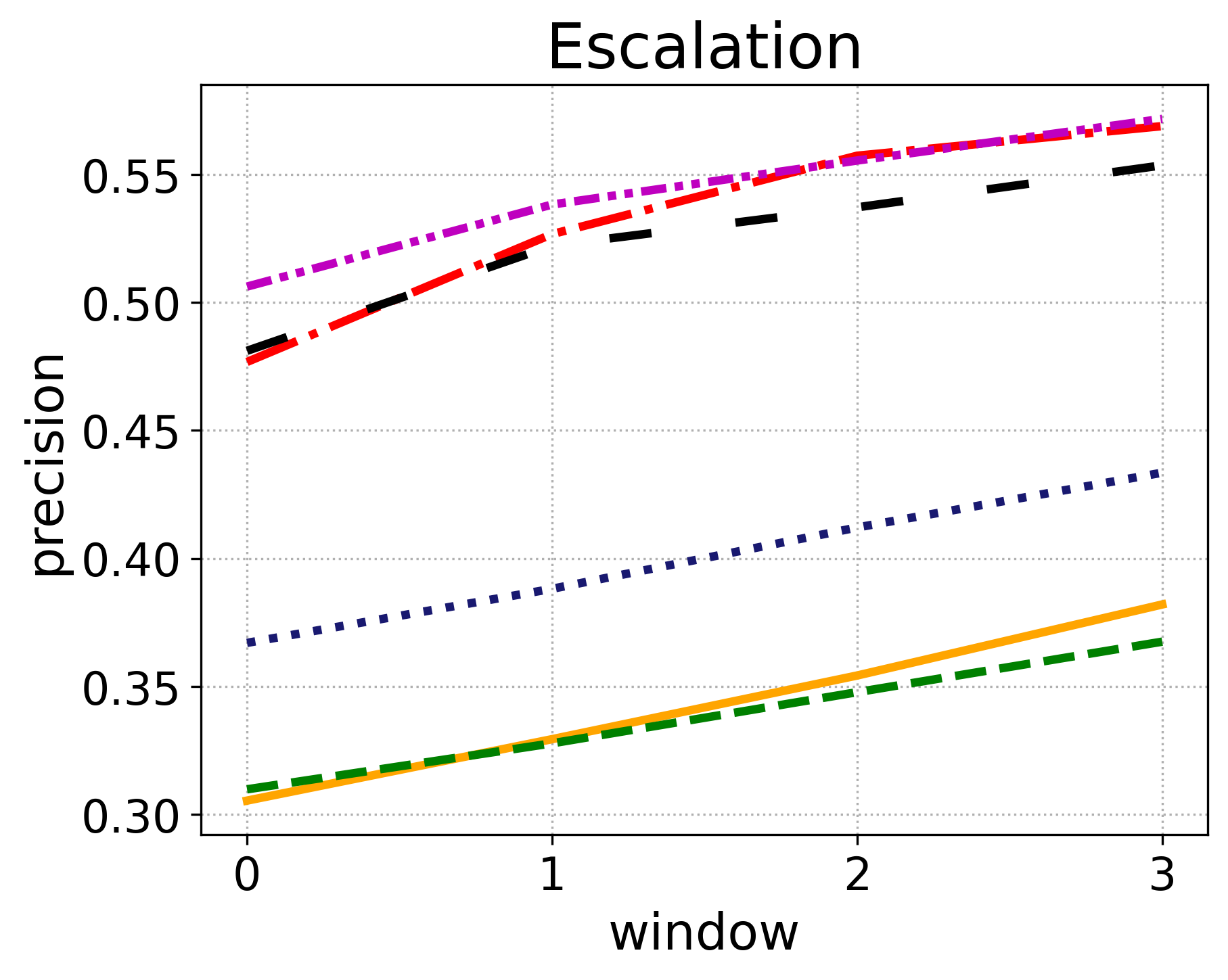}  
  \label{fig:sub-second}
\end{subfigure}
\begin{subfigure}{.245\linewidth}
  \centering
  \includegraphics[width=\linewidth]{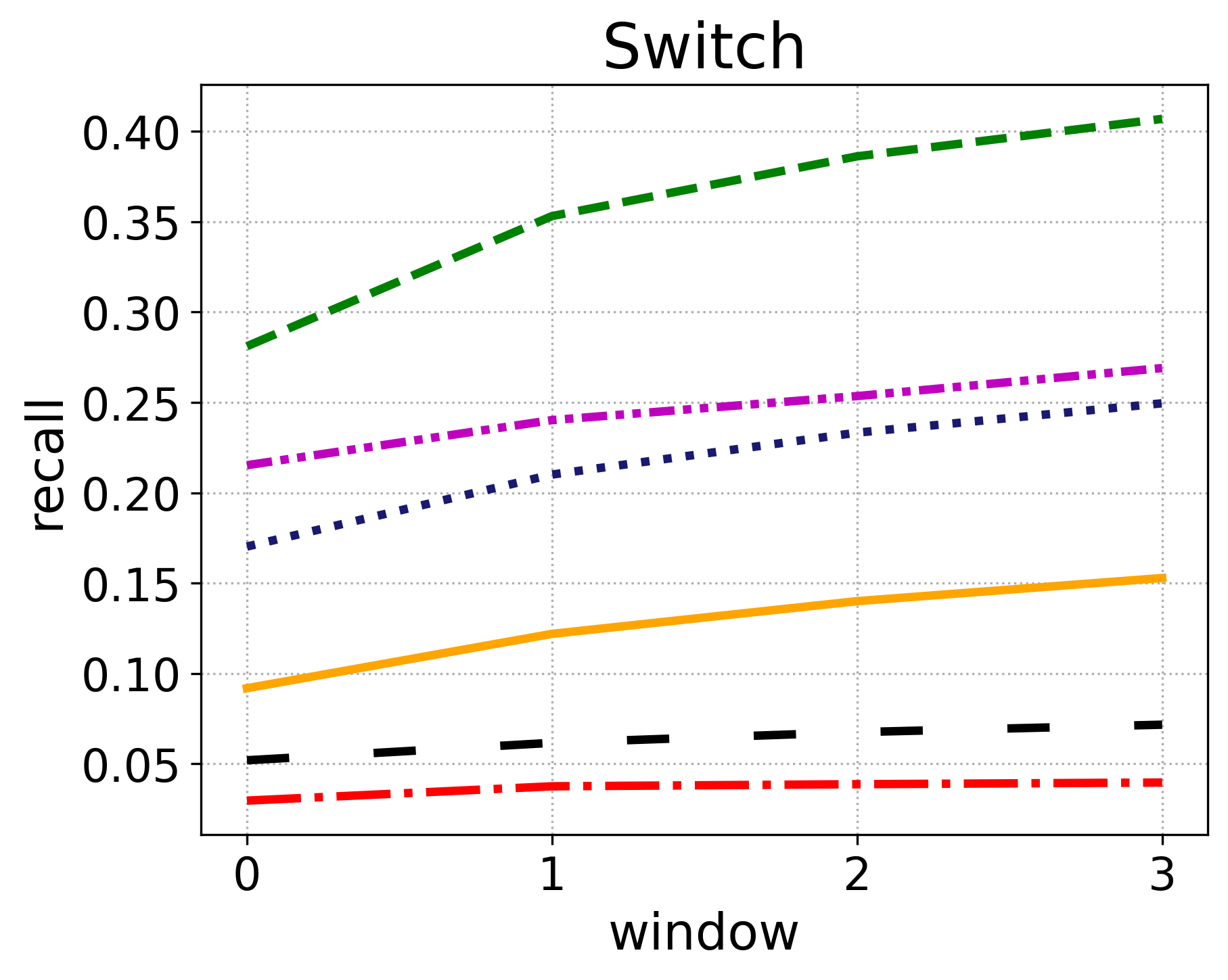}  
  \label{fig:sub-second}
\end{subfigure}
\begin{subfigure}{.245\linewidth}
  \centering
  \includegraphics[width=\linewidth]{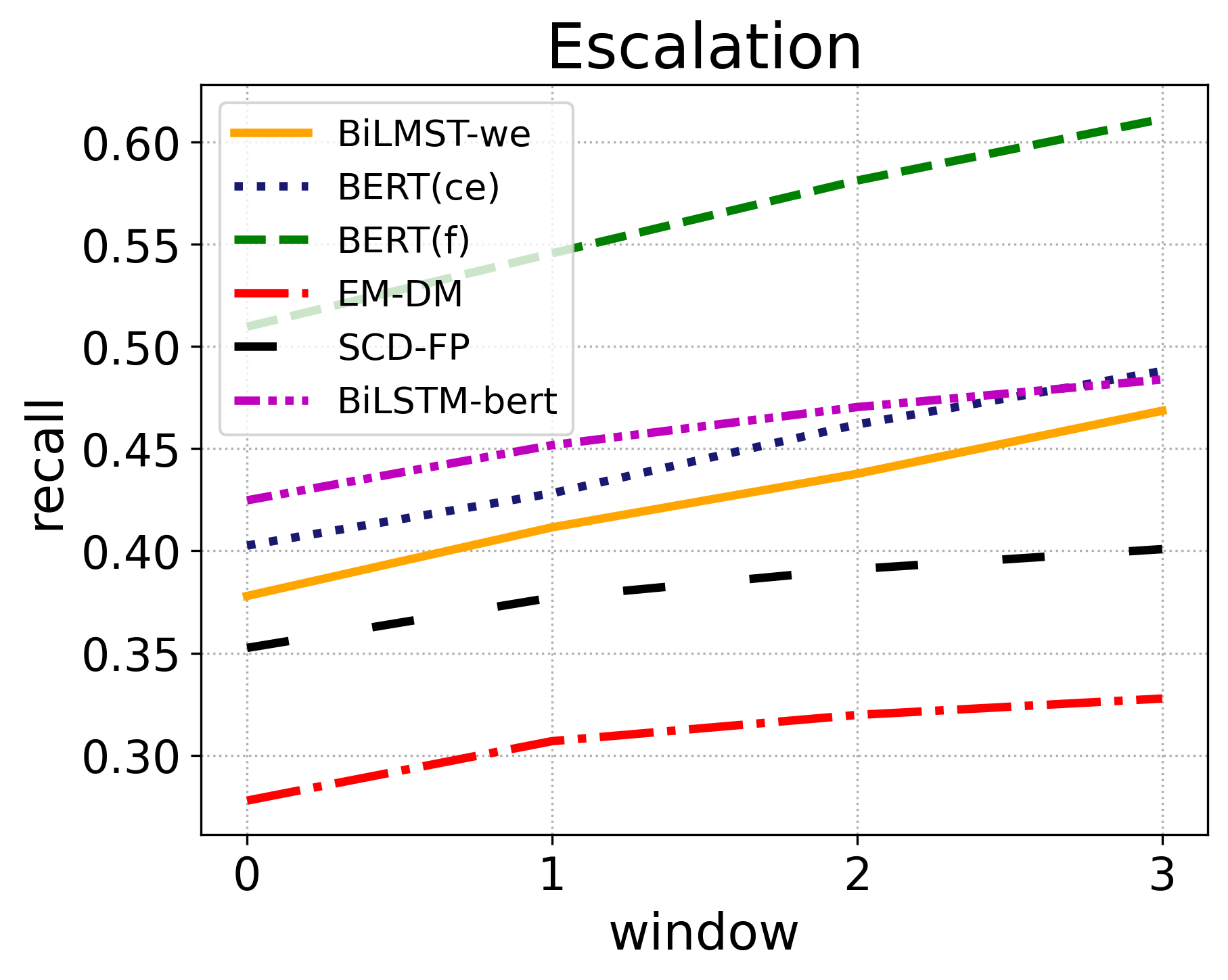}  
  \label{fig:sub-second}
\end{subfigure}
\caption{Timeline-level Precision $P_w$ and Recall $R_w$ of the best performing models.}
\label{fig:longitudinaleval}
\end{figure*}

\section{Results \& Discussion}

\subsection{Quantitative Comparison}\label{sec:resultsquantitative}

\paragraph{Model Comparison} Table~\ref{tab:postlevelresults} summarises the results of all models; Fig.~\ref{fig:longitudinaleval} further shows the $P_w$/$R_w$ metrics for IE/IS for the best-performing models. \texttt{BiLSTM-bert} confidently outperforms all competing models in terms of post-level macro-F1. It provides a 8.6\% relative improvement (14\% for the IS/IE labels) against the second best performing model (\texttt{BERT(f)}). Furthermore, it achieves a great balance between precision- and recall-oriented timeline-level metrics, being consistently the second-best performing model. This performance is largely attributed to two factors, which are studied further below: (a) the use of the Focal loss on BERT, generating \texttt{[CLS]} representations that are much more focused on the minority classes (IE/IS), and (b) its longitudinal aspect. 

\paragraph{Post-level} The BERT variants perform better than the rest in all metrics. Their coverage metrics though suggest that while they manage to predict better the regions compared to most timeline-level methods (i.e., high $C_r$), they tend to predict more regions than needed (i.e., low $C_p$) -- partially due to their lack of contextual (temporal-wise) information. Finally, as expected, \texttt{BERT(f)} achieves much higher recall for the minority classes (IE/IS), in exchange for a drop in precision compared to \texttt{BERT(ce)} and in recall for the majority class (O).

\paragraph{Models from Related Tasks} \texttt{EM-DM} achieves very high precision ($P$, $P_w$) for the minority classes, showing a clear link between the tasks of emotion recognition and detecting changes in a user's mood -- indeed, emotionally informed models have been successfully applied to post-level classification tasks in mental health~\cite{sawhney-etal-2020-deep}; however, both \texttt{EM} models achieve low recall ($R$, $R_w$) for IE/IS compared to the rest. For the SCD inspired models, \texttt{SCD-FP} outperforms \texttt{SCD-OP} on most metrics. This is largely due to the fact that the former uses the previous $k$=3 posts to predict the next post in a user's timeline (instead of aligning it based on the previous post only. 
 Thus \texttt{SCD-FP} benefits from its longitudinal component -- a finding consistent with work in semantic change detection~\cite{tsakalidis-liakata-2020-sequential}.

\paragraph{Representation \textit{vs} Fine-tuning \textit{vs} Focal Loss} While \texttt{BiLSTM-bert} yields the highest macro-F1 and the most robust performance across all metrics, it is not clear which of its components contributes the most to our task. 
  To answer this, we perform a comparison against the exact same BiLSTM, albeit fed with different input types: (a) average word embeddings as in \texttt{BiLSTM-we}, (b) Sentence-BERT representations \cite{reimers2019sentence} and (c) fine-tuned representations from \texttt{BERT(ce)}. As shown in Table~\ref{tab:postlevelresultsbilstm}, fine-tuning with \texttt{BERT(ce)} outperforms Sentence-BERT representations. While the contextual nature of all of the BERT-based models offers a clear improvement over the static word embeddings, it becomes evident that the use of the focal loss during training the initial \texttt{BERT(f)} is vital, offering a  relative improvement of 6\% in post-level macro-F1 (13.7\% for IS/IE). Calibrating the parameters in the focal loss could provide further improvements for our task in the future \cite{mukhoti2020calibrating}.

\begin{table}[]
\centering
\resizebox{.95\linewidth}{!}{%
\begin{tabular}{l|rrr|rr|rr|}
 & \multicolumn{3}{c}{\textbf{Post}} & \multicolumn{2}{c}{\textbf{Timeline}} & \multicolumn{2}{c|}{\textbf{Coverage}} \\\cline{2-8}
 & P & R & F1 & P$_1$ & R$_1$  & $C_p$&$C_r$ \\\cline{2-8}
Word emb. & .589 & .488 & .508 &.577&.450&.412&.282\\
Sent.-BERT & .610 & .535 & .546 &.601&.499&.428&.333\\
BERT(ce) & .612 & .518 & .554 &\textbf{.624}&.520&.434&.378\\
BERT(f) & \textbf{.621} & \textbf{.553} & \textbf{.580} &.622&\textbf{.545}&\textbf{.447}&\textbf{.398}
\end{tabular}%
}
\caption{Macro-avg performance of timeline-level BiLSTM operating on different input representations (see \textbf{Representation \textit{vs} Fine-tuning \textit{vs} Focal Loss} in \S\ref{sec:resultsquantitative}).}
\label{tab:postlevelresultsbilstm}
\end{table}

\begin{figure}
    \centering
    \includegraphics[width=.95\linewidth]{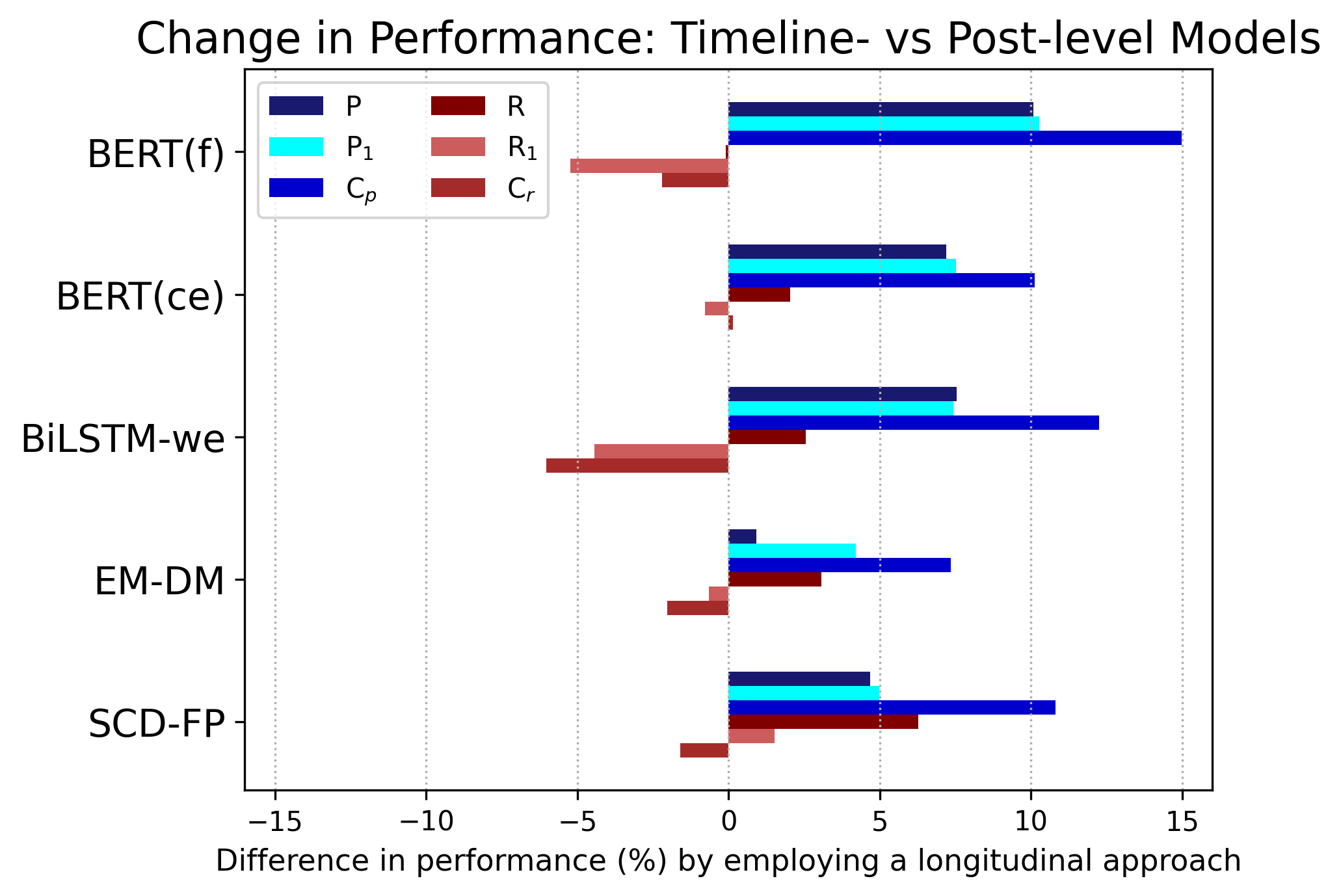}
    \caption{Gains/losses in performance (\%) when incorporating a longitudinal component for each model (see \textbf{Timeline- \textit{vs} Post-level Modelling} in \S\ref{sec:resultsquantitative}).}
    \label{fig:longitudinalcomparison}
\end{figure}

\paragraph{Timeline- \textit{vs} Post-level Modelling} The importance of longitudinal modelling is shown via the difference between the BERT and BiLSTM variants when operating on single posts \textit{vs} on the timeline-level (e.g., see the post-level results of \texttt{BERT(ce)}/\texttt{Word emb.} in Table~\ref{tab:postlevelresultsbilstm} \textit{vs} \texttt{BERT(ce)}/\texttt{BiLSTM-we} in Table~\ref{tab:postlevelresults}, respectively). We further examine the role of longitudinal modelling in the rest of our best-performing models from Table~\ref{tab:postlevelresults}. In particular, we replace the timeline-level BiLSTM in \texttt{EM-DM} and \texttt{SCD-FP} with a two-layer feed-forward network, operating on post-level input representations -- treating each post in isolation. The differences across all pairwise combinations with and without the longitudinal component are shown in Fig.~\ref{fig:longitudinalcomparison}. Timeline-level models achieve much higher precision (6.1\%/6.9\%/11.1\% for $P$/$P_1$/$C_p$, respectively) in return for a small sacrifice in the timeline-level recall-oriented metrics (-2.8\%/1.9\%/2.3\% for $R$/$R_1$/$C_r$), further highlighting the longitudinal nature of the task. 

\subsection{Qualitative Analysis}\label{sec:qualitative}
Here we 
analyse the cases of Switches/Escalations identified or missed by our best performing model (\texttt{BiLSTM-bert}).

\begin{figure}
    \centering
    \includegraphics[width=.75\linewidth]{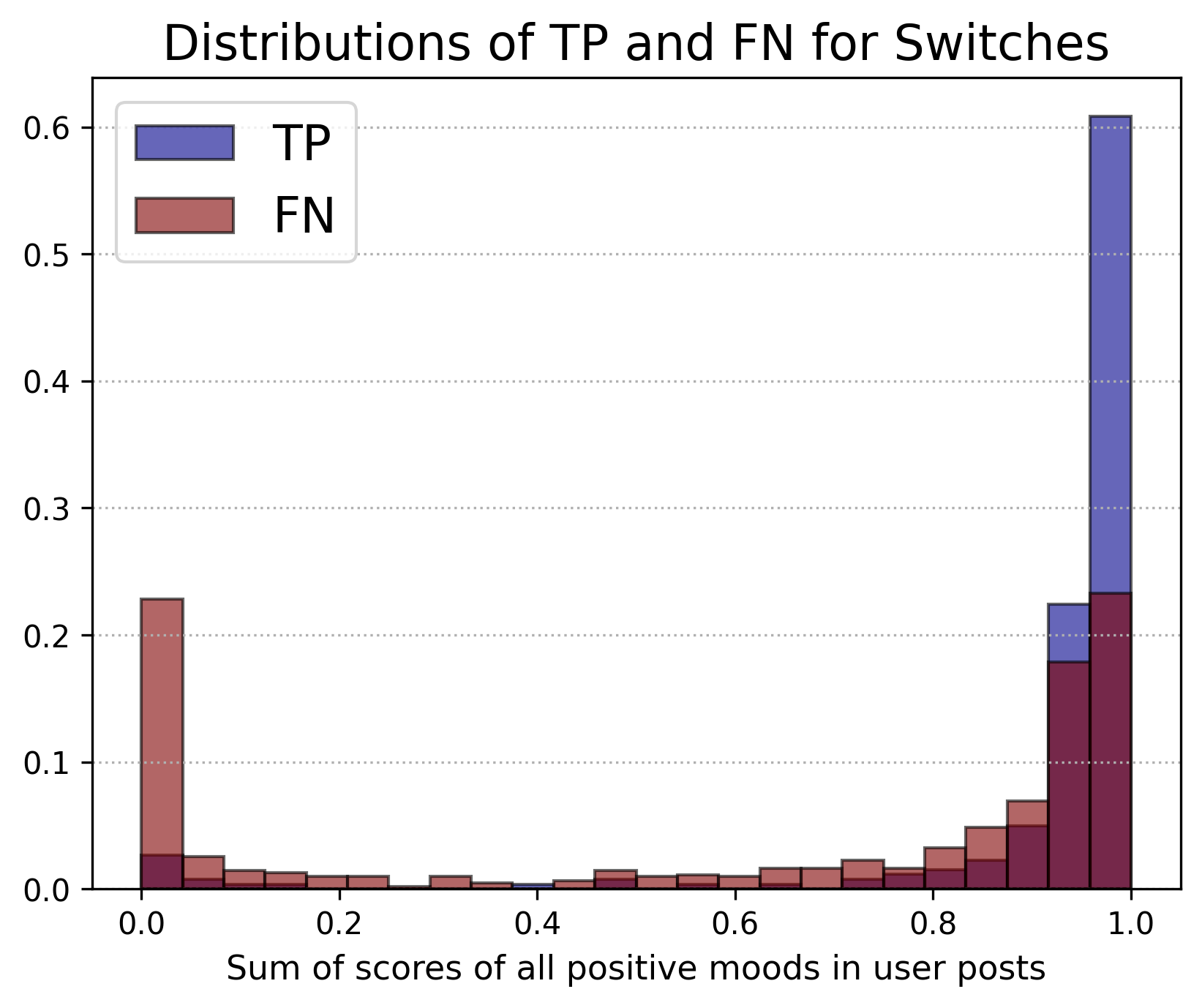}
    \caption{Histogram of positive emotion scores in True Positive \& False Negative distributions, for the Switch label.}
    \label{fig:switch_analysis}
\end{figure}

\vspace{.1cm}
\noindent\textbf{Switches (IS)} are the most challenging to identify, largely due to being the smallest class with the lowest inter-annotator agreement. 
 However, the \texttt{EM}-based models achieve high levels of precision on Switches, even during post-level evaluation (see Table~\ref{tab:postlevelresults}). We therefore employ \texttt{EM-TR} \cite{barbieri-etal-2020-tweeteval}, assigning probability scores for anger/joy/optimism/sadness to each post, and use them to characterise the predictions made by 
\texttt{BiLSTM-bert}. 
Fig.~\ref{fig:switch_analysis} and Table~\ref{tab:switch_analysis} show that our model predicts more often (in most cases, correctly) a `Switch' when the associated posts express positive emotions (joy/optimism), but misses the vast majority of cases when these emotions are absent. The reason for this is that TalkLife users discuss issues around their well-being, with a negative mood prevailing. Therefore, \texttt{BiLSTM-bert} learns that the negative tone forms the users' baseline and thus deviations from this constitute cases of `Switches' (see example in Table~\ref{tab:switch_analysis_example}). We plan to address this in the future by incorporating transfer learning approaches to our model~\cite{ruder2019transfer}.

\begin{table}[]
\centering
\resizebox{.67\linewidth}{!}{%
\begin{tabular}{lrrrr}
&\textbf{angry}&\textbf{joy}&\textbf{optim.}&\textbf{sad.}\\ \hline
\textbf{TP} &.03& .76& .14& .07\\
\textbf{FP} &.06& .60& .19& .15\\
\textbf{FN} &.13& .44& .18& .25
\end{tabular}%
}
\caption{Average probability of each emotion per classification case on `Switches' (see \textbf{Switches} in \S\ref{sec:qualitative}).}
\label{tab:switch_analysis}
\end{table}

\begin{table}[]
\resizebox{\linewidth}{!}{%
\begin{tabular}{lcc}
\textbf{Text} & \textbf{True} & \textbf{Pred.} \\\hline\hline
Oh, forgot :) Stay safe you lovely people all around&&\\the world! & O & IS \\\hline
Hope you are all having a good night! Stay safe! :D & O & IS \\\hline
Don't wanna deal with anyone.. Hope school finishes&&\\so I can go home soon & IS & O \\\hline
Tired of my leg hurting so badly today. I really can't&&\\do any training :( & IS & O \\\hline
Hope you're all great! <3 Love you all! & O & IS
\end{tabular}%
}
\caption{Example of a Switch in part of a user's (paraphrased) timeline, missed by \texttt{BiLSTM-bert}. 
}
\label{tab:switch_analysis_example}
\end{table}

\begin{figure}
    \centering
    \includegraphics[width=.8\linewidth]{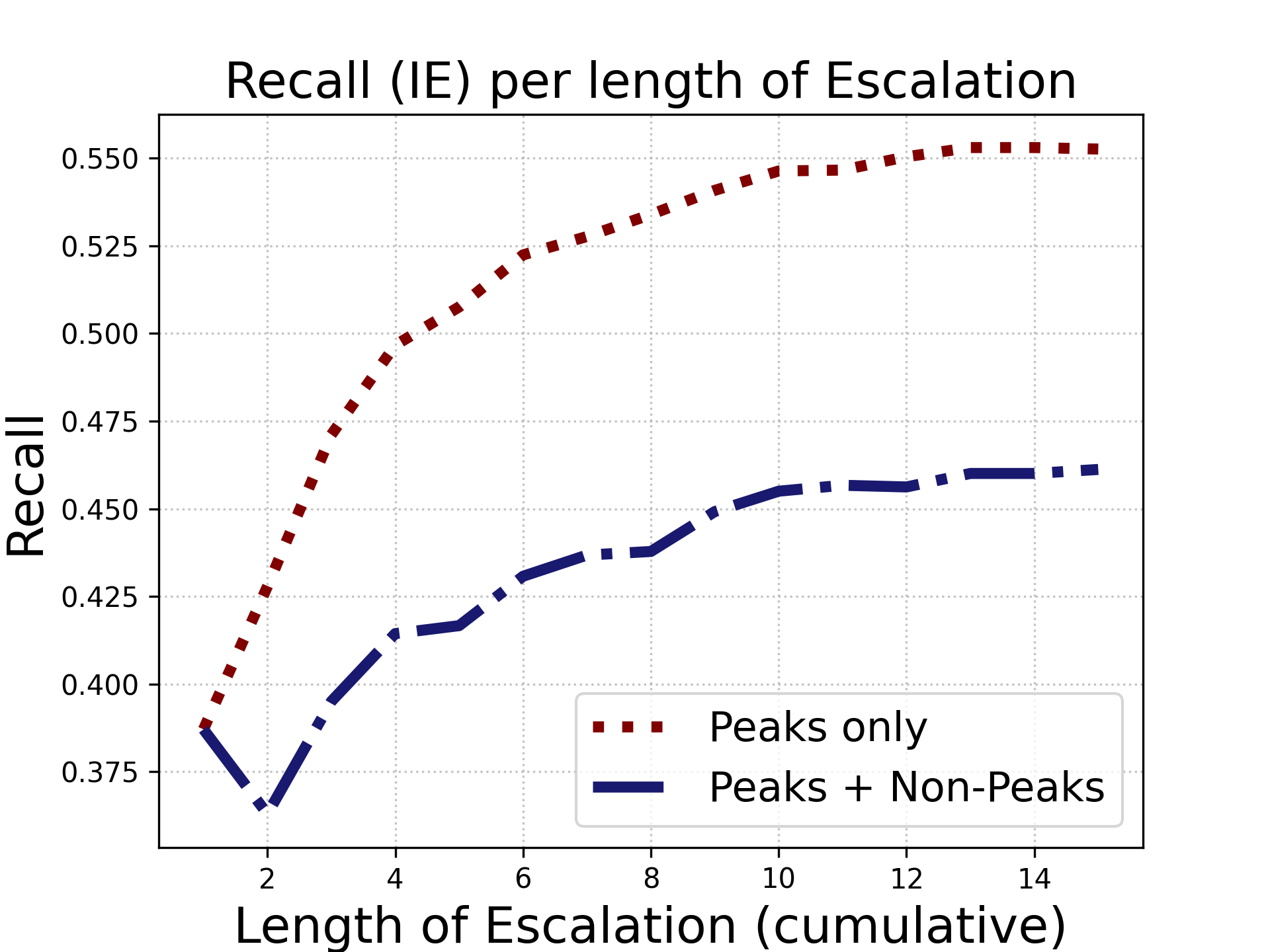}
    \caption{Recall for IE cases per cumulative length of Escalation (see \textbf{Escalations} in \S\ref{sec:qualitative}).}
    \label{fig:escalation_analysis}
\end{figure}

\vspace{.2cm}
\noindent\textbf{Escalations (IE)} are better captured by our models. Here we examine more closely the cases of `Peaks' in the escalations (i.e., the posts indicating the most negative/positive state of the user within an escalation -- see \S\ref{sec:annotationtool}). As expected, the post-level recall of \texttt{BiLSTM-bert} in these cases is much higher than its recall for the rest of IE cases (.557 vs .408). In Fig.~\ref{fig:escalation_analysis} we analyse the recall of our model in capturing posts denoting escalations, in relation to the length of escalations. 
We can see that our model is more effective in capturing longer escalations. As opposed to the Switch class, we found no important differences in the expressed emotion between TP and FN cases. By carefully examining the cases of Peaks in isolation, we found that the majority of them express very negative emotions, very often including indication of self-harm. A Logistic Regression trained on bigrams at the post-level to distinguish between identified \textit{vs} missed cases of Peaks showed that the most positively correlated features for the identified cases were directly linked to self-harm (e.g., \textit{``kill myself''}, \textit{``to die''}, \textit{``kill me''}). However, this was not necessarily the case with missed cases. 
Nevertheless, there were several cases of self-harm ideation that were missed by \texttt{BiLSTM-bert}, as well as misses due to the model ``ignoring'' the user's baseline, as is the case with Switches (see Table~\ref{tab:escalation_analysis_example}). Transfer learning and domain adaptation strategies as well as self-harm detection models operating at the post level could help in mitigating this problem. 

\begin{table}[]
\centering
\resizebox{\linewidth}{!}{%
\begin{tabular}{l}
\textbf{Text} \\\hline\hline
When my parents go out, I am gonna cut. \\ \hline
I feel so horrible. I really don't want to be here anymore.\\\hline
Someone please text me... I swear I am about to harm myself...\\Please, anyone!'\\\hline
Had an awesome day with my gf and she tagged me! I am not\\alone! :)\\\hline
Have not cut for the past year!! Yay!!
\end{tabular}%
}
\caption{Examples of Peaks of Escalations (isolated paraphrased posts) missed by \texttt{BiLSTM-bert}.}
\label{tab:escalation_analysis_example}
\end{table}

%% file: 6conclusion.tex
\section{Conclusion and Future Work}

We present a novel longitudinal dataset and associated models for personalised monitoring of a user's well-being over time based on linguistic online content. Our dataset contains annotations for: (a) sudden shifts in a user's mood (switches) and (b) gradual mood progression (escalations). Proposed methods are inspired by state-of-the-art contextual models and longitudinal NLP tasks. Importantly, we have introduced temporally sensitive evaluation metrics, adapted from the fields of change-point detection and image segmentation. Our results highlight the importance of considering the temporal aspect of the task and the rarity of mood changes.\smallskip

\noindent \textbf{Future work} could follow four main directions: (a) integrating longitudinal models of detecting changes, with post-level models for emotion and self-harm detection 
(see \S\ref{sec:qualitative}); 
(b) incorporating transfer learning methods \cite{ruder2019transfer} to adapt more effectively to unseen users' timelines; (c) adjusting our models to learn from multiple (noisy) annotators~\cite{paun2021aggregating} 
 and (d) calibrating the parameters of focal loss and testing other loss functions suited 
to heavily imbalanced classification tasks \cite{jadon2020survey}.

%% file: 7appendix.tex
\appendix

\section{Hyperparameters}
\label{sec:appendix}

Here we provide details on the hyperparameters used by each of our models, presented in \S\ref{sec:approaches}:

\begin{itemize}
    \item \texttt{RF}: Number of trees: [50, 100, 250, 500] 
    \item \texttt{BiLSTM-we}: Two hidden layers ([64,128,256] units), each followed by a drop-out layer (rate: [.25, .5, .75]) and a final dense layer for the prediction. Trained for 100 epochs (early stopping if no improvement on 5 consecutive epochs) using Adam optimizer (lr: [0.001, 0.0001]) optimzing the Cross-Entropy loss with batches of size [128, 256], limited to modelling the first 35 words of each post. 
    \item \texttt{BiLSTM-bert}: Two hidden layers ([64,128,256] and [124] units, respectively), each followed by a drop-out layer (rate: [.25, .5, .75]) and a final dense layer on each timestep for the prediction. Trained for 100 epochs (early stopping if no improvement on 5 consecutive epochs) using Adam optimizer \cite{kingma2014method} (lr: [0.001, 0.0001]) optimizing the Cross-Entropy loss with batches of size [16, 32, 64].
    \item \texttt{EM-DM} \& \texttt{EM-TR}: Same architecture as \texttt{BiLSTM-bert}, albeit operating on the \texttt{EM-DM}'s (\texttt{EM-TR}'s)  output.
    \item \texttt{FSD}: Same architecture as \texttt{BiLSTM-bert}. For the FSD part, we experimented with word embeddings\footnote{en-core-web-lg @ \url{https://github.com/explosion/spacy-models/releases/download/en_core_web_lg-3.0.0/en_core_web_lg-3.0.0-py3-none-any.whl}} and representations from Sentence-BERT. We extract features either by considering the nearest neighbor or by considering the centroid, on the basis of the previous [1,2,...,10] posts, as well as on the basis of the complete timeline preceding the current post (11 features, overall). The two versions (nearest neighbor, centroid) were run independently from each other.
    \item \texttt{SCD-OP} \& \texttt{SCD-FP}: We experimented with average post-level word embeddings and representations from Sentence-BERT (results are reported for the latter, as it performed better). For \texttt{SCD-FP}, we stacked two BiLSTM layers (128 units each), each followed by a dropout (rate: 0.25), and a final dense layer for the prediction, with its size being the same as the desired output size (300 for the case of word embeddings, 768 for Sentence-BERT). We train in batches of 64, optimising the cosine similarity via the Adam Optimizer with a learning rate of .0001, and employing an early stopping criterion (5 epochs patience). The final model (i.e., after the SCD part) follows the exact same specifications as \texttt{BiLSTM-bert}, operating on the outputs from the SCD components.
    \item \texttt{BERT(ce)} \& \texttt{BERT(f)}: We used BERT-base (uncased) as our base model and added a Dropout layer (rate: .25) operating on top of the \texttt{[CLS]} output, followed by a linear layer for the class prediction.  We trained our models for 3 epochs using Adam (learning rate: [1e-5, 3e-5]) and perform five runs with different random seeds (0, 1, 12, 123, 1234). Batch sizes of 8 are used in train/dev/test sets. For the alpha-weighted Focal loss in \texttt{BERT(f)}, we used $\gamma=2$ and $a_t=\sqrt{1/p_t}$, where $p_t$ is the probability of class $t$ in our training data. Results reported in the paper (as well as the results for BiLSTM-bert) are averaged across the five runs with the different random seeds.
\end{itemize}

We trained each model on five folds and selected the best-performing combination of hyperparameters on the basis of macro-F1 on a dev set (33\% of training data) for each test fold.

\section{Libraries}

The code for the experiments is written in Python 3.8 and relies on the following libraries: keras (2.7.0), numpy (1.19.5), pandas (1.2.3), scikit-learn (1.0.1), sentence\_trasformers (1.1.0), spacy (3.0.5), tensorflow (2.5.0), torch (1.8.1), transformers (4.5.1).

\section{Infrastructure}

All experiments were conducted on virtual machines (VM) deployed on the cloud computing platform Microsoft Azure. We have used two different VMs in our work:
\begin{itemize}
    \item the experiments that involved the use of BERT were ran on a Standard NC12\_Promo, with 12 cpus, 112 GiB of RAM and 2 GPUs;
    \item all other experiments were ran on a Standard F16s\_v2, with 16 cpus and 32 GiB of RAM.
\end{itemize}